# Exploring Agentic Artificial Intelligence Systems: Towards a Typological Framework

*Completed Research Paper*

**Christopher Wissuchek**
Leipzig University
Leipzig, Germany
christopher.wissuchek@uni-leipzig.de

**Patrick Zschech**
TU Dresden
Dresden, Germany
patrick.zschech@tu-dresden.de

## Abstract

*Artificial intelligence (AI) systems are evolving beyond passive tools into autonomous agents capable of reasoning, adapting, and acting with minimal human intervention. Despite their growing presence, a structured framework is lacking to classify and compare these systems. This paper develops a typology of agentic AI systems, introducing eight dimensions that define their cognitive and environmental agency in an ordinal structure. Using a multi-phase methodological approach, we construct and refine this typology, which is then evaluated through a human-AI hybrid approach and further distilled into constructed types. The framework enables researchers and practitioners to analyze varying levels of agency in AI systems. By offering a structured perspective on the progression of AI capabilities, the typology provides a foundation for assessing current systems and anticipating future developments in agentic AI.*

**Keywords:** Agentic Artificial Intelligence, AI Agents, Typology

## Introduction

Artificial intelligence (AI) is increasingly being adopted, with organizations allocating substantial resources toward its implementation. AI systems, which perform cognitive functions, offer significant opportunities (Deng, 2018; Rai et al., 2019). These capabilities have positioned AI as a critical component of contemporary organizational strategies, contributing to competitive advantages and becoming a general-purpose technology supporting various applications (Krakowski et al., 2023). Initially, AI systems were primarily considered analytical tools for structured and routine tasks optimized for specific, well-defined, or rule-based decision-making scenarios (Feuerriegel et al., 2024).

Large models, such as large language models (LLMs), represent a significant advancement and substantially expand AI's scope (Schneider et al., 2024). Unlike traditional AI systems reliant on mostly discriminative approaches, which learn the decision boundary between pre-defined classes, they employ transformer models to create content, handle complex reasoning tasks, and adapt dynamically to various contexts (Feuerriegel et al., 2024). This evolution broadens AI's applicability, particularly in knowledge-intensive domains where tasks require high adaptability, creativity, or contextual understanding. AI is becoming gradually more pervasive in handling tasks previously reserved for humans that were considered challenging to automate (Acharya et al., 2025; Riemer & Peter, 2024).

In this context, AI systems are increasingly capable of making or recommending decisions on behalf of human agents, prompting a reevaluation of the human-technology relationship. From an information systems (IS) perspective, where artifacts have traditionally been regarded as passive tools acted upon by human users (Baird & Maruping, 2021), AI represents a shift in agency, extending it to systems that operate with minimal human intervention by initiating actions, dynamically adapting to changing contexts, and independently setting and pursuing cascading goals (Acharya et al., 2025).

An example of an agentic AI system is GitHub Copilot, which autonomously observes the software engineer, monitors the development environment, tracks changes, and adaptively generates or modifies code (Cui et al., 2024). In contrast, a non-agentic AI system, such as an analytical tool, provides insights from structured data only when explicitly initiated by a human, who retains complete control over the outcome (Wissuchek & Zschech, 2024). These examples illustrate that agency in AI exists on a spectrum shaped by factors such as human involvement, the system's scope of influence, goal complexity, time horizon, and responsiveness to change. At one end, passive systems act only on direct input; reactive systems respond to stimuli with pre-defined actions. More advanced systems may anticipate needs and act without prompts, while autonomous agents operate independently (Acharya et al., 2025; Baird & Maruping, 2021).

Given recent advancements and the significant business potential of agentic AI systems, the industry is rapidly advancing toward their adoption. This progress is fueled by the availability of open-source and commercially developed foundation models, which are integrated into existing IT platforms or leveraged to create novel applications (Schneider et al., 2024). These systems also enable organizations to build tailored solutions, often utilizing extendable, off-the-shelf agents that require minimal technical expertise (Speer et al., 2024). While the current enthusiasm may reflect the peak of a hype cycle, it is plausible that these systems will continue to evolve and advance rapidly, driven by growing investments both from the private and public sectors, such as the $500 billion commitment to the Stargate Project (OpenAI, 2025a).

Despite these advancements, there are challenges in fully understanding, categorizing, and effectively deploying the rapidly evolving capabilities of agentic AI systems – particularly regarding their ability to independently initiate actions, adapt to dynamic environments, and pursue goals autonomously (Göldi & Rietsche, 2024; Murugesan, 2025). These challenges are further compounded by potential unforeseen consequences of increasingly agentic behavior (Shavit et al., 2023). The variability in how agency manifests across AI systems makes it difficult for organizations to accurately assess their suitability for specific tasks and strategically integrate them into operational structures and workflows (Krakowski, 2025). Without a systematic framework, we argue that organizations risk misalignment between agentic AI capabilities and practical applications, potentially undermining effective implementation and long-term planning.

A widely used strategy for managing such complexities involves developing structured classifications. Typology-based frameworks are common approaches to making sense of existing developments and anticipating future advancements in a field (Bailey, 1994). Such frameworks can serve as reference points for phenomena or provide a robust theoretical foundation for future research (Negoita et al., 2018). Recent literature has begun exploring the complexity of agentic AI by providing valuable classifications and overview studies. These include technical analyses, domain-specific examinations, and assessments of subsets of agentic AI systems (e.g., Acharya et al., 2025; Durante et al., 2024; Zhao et al., 2023). Despite these contributions, there remains a gap in establishing a comprehensive framework that systematically conceptualizes the core aspects of agentic behavior in AI, independent of specific implementations or modeling approaches. Such a framework would enable the identification of distinct characteristics and variations while also providing a foundation for anticipating future developments and impacts. To address this gap, we strive to achieve the following research objective:

> *Build a typological framework to characterize and enable the exploration of agentic AI systems.*

By applying a typological framework-based approach, this paper contributes to and advances the understanding of agentic AI systems in several ways. *First*, it introduces a new and precise definition of agentic AI systems. *Second*, it develops a typology for their classification, enabling comparisons between non-agentic and agentic AI systems as well as different variations, supporting researchers and practitioners in making more informed decisions or evaluations. *Third*, the paper serves as a launchpad, offering a foundation for research or theorizing on application scenarios, consequences, design science artifacts, or propositions, among other potential directions.

The remainder of this paper is structured as follows: First, we present the theoretical background, focusing specifically on agency in AI. Subsequently, we detail our methodological approach to developing the typology, encompassing six distinct phases. This is followed by a section demonstrating the practical application and utility of the typology using empirical examples. Finally, we discuss our findings, address the study's limitations, and suggest directions for future research.

# Background

## *Agency, Agents, and "Agenticness"*

Agency can be broadly described as the capacity of an entity to act independently, make decisions, and pursue objectives, characterizing it as an active participant rather than a passive recipient of external influences (Schlosser, 2019). Agency is often attributed to humans, emphasizing autonomy, purposeful behavior, and the ability to shape their environment. While agency refers to this intrinsic capacity, an agent is an entity demonstrating this capability (Schlosser, 2019). Additionally, following Shavit et al. (2023), we use "agenticness" to describe agency along a spectrum rather than as a binary property.

The term "agent" has a long history in IS research and related fields like computer science, where it is often used loosely to describe software systems exhibiting – or appearing to exhibit – intelligent behavior (Decker & Sycara, 1997; Rudowsky, 2004). Common examples include conversational agents, personal assistants, chatbots that interact using natural language, and smart or sensing systems such as predictive maintenance (Leitao et al., 2016). These agents are typically distinguished by their communication mode, embodiment, and application context (Diederich et al., 2022). While the term frequently applies to systems that mimic human-like behavior or appearance, its use is often metaphorical, focusing on perceived characteristics rather than foundational agency. These systems generally function as artifacts that resemble human behavior but do not inherently possess agency. Nonetheless, this framing helps conceptualize them as social actors and supports their integration into human-centered environments (Göldi & Rietsche, 2024). In response to the term's ambiguity, recent literature increasingly favors "agentic" over "agent" when referring to systems in the context of agency (e.g., Acharya et al., 2025; Baird & Maruping, 2021; Shavit et al., 2023).

Against this backdrop, the question arises: what does it mean to be agentic? While extensively debated, resulting in various conceptions and theories (Schlosser, 2019), we aim to narrow the concept to within our research's scope. A central aspect of agency is the capacity to act and effect change, often through intentional actions. In standard theory, such actions are driven by mental representations – desires, beliefs, or emotions – and guided by reasons and rational deliberation, making agency inherently human-centric (Schlosser, 2019). However, whether human-engineered agents, such as software systems, can ever achieve full-fledged agency remains highly debated (Popa, 2021). Although serious discussions exist on the possibility that artificial agents may develop mental representations that result in intentions, these discussions remain speculative. A significant shortcoming of the standard theory is its dependence on mental states despite observable agent-like behavior in non-human entities such as AI systems (Popa, 2021). Therefore, we adopt a more restrictive, instrumentalist viewpoint that does not require mental representations (Popa, 2021), allowing us to assess the functionality of artificial agents – even when their actions are simulated and guided by human design rather than arising naturally.

One such view is offered by Floridi and Sanders (2004), who outline three essential features for an artificial entity to be considered agentic: (1) Interactivity, the capacity to engage with the environment by both affecting and being affected by external conditions; (2) Autonomy, the ability to function without direct external control; and (3) Adaptability, the capability to learn from past interactions and refine behavior over time. This perspective does not include goal-directedness, even though artificial systems are typically designed to fulfill specific objectives. In contrast, Barandiaran et al. (2009) propose a concept of minimal agency defined by three conditions: (1) Individuality, the agent exists as a distinct entity separate from its environment; (2) Interactional asymmetry, it can independently initiate actions within its environment; and (3) Normativity, its behavior is guided by specific goals or objectives. There is a clear overlap between these two perspectives, particularly regarding interactivity and autonomy. Unlike Floridi and Sanders' model (2004), minimal agency explicitly incorporates goal-directed behavior. In summary, for our exploration of agency in AI, we integrate these perspectives and define four conditions of agenticness: (1) Interactivity, (2) Autonomy, (3) Adaptability, and (4) Normitivity.

## *Agency in the Context of AI*

AI systems are increasingly perceived as exhibiting agency, particularly as they take on tasks traditionally associated with human intelligence – such as decision-making, problem-solving, reasoning, planning, knowledge representation, content generation, and perception (Popa, 2021; Rai et al., 2019). This shift is closely tied to recent advances in machine learning, especially model advancements, the availability of large datasets, and increased computational power, which has enabled the development of more agentic AI systems (Deng, 2018; Murugesan, 2025). In the following, we will discuss agency in the context of some of the currently dominant AI paradigms.

Supervised learning, a dominant machine learning paradigm, primarily relies on discriminative modeling to distinguish between pre-defined categories. It maps input features to specific outputs using labeled datasets, enabling functions such as classification and prediction. While supervised learning facilitates automation and enables AI systems to make data-driven decisions, its behavior is limited by the structure of its training data (Janiesch et al., 2021). These models do not operate beyond their pre-defined labels and cannot dynamically adapt their objectives (Jiang et al., 2020). However, supervised learning-based AI systems may exhibit interactivity by processing and responding to inputs and a limited form of autonomy, as they can potentially execute tasks independently once trained (Langer & Landers, 2021).

Reinforcement learning (RL) offers a more dynamic approach than discriminative methods. RL employs an agent that interacts with its environment, learning through trial and error based on rewards or penalties, including human feedback. This iterative process enables agents to optimize decision-making strategies for long-term rewards rather than immediate classifications (Sutton & Barto, 2018). RL satisfies interactivity and adaptability as agents refine behavior in response to environmental changes. It can also exhibit limited autonomy, with agents independently selecting actions within a defined framework (Ladosz et al., 2022). Although goals remain externally set by humans, RL allows AI to explore actions, assess outcomes, and adjust behavior – representing a meaningful step toward agenticness beyond supervised learning.

Large model-based approaches mark a significant advancement, integrating multiple modeling techniques (Y. Wang et al., 2024). Unlike models built for specific tasks, they rely on task-agnostic foundation models adaptable to various applications. Trained on vast datasets using self-supervised learning, these models capture complex patterns without explicit task-specific programming (Schneider et al., 2024; Zhao et al., 2023). They generalize knowledge across domains, enabling the generation of novel outputs like text, images, and code. At their core are deep neural networks that learn probabilistic relationships rather than storing explicit examples, allowing content generation based on underlying structures (Banh & Strobel, 2023; Feuerriegel et al., 2024). LLMs exemplify this, using transformer architectures to interpret and produce text by evaluating contextual word relationships (Zhao et al., 2023). Reinforcement learning from human feedback further fine-tunes these models, enhancing coherence and alignment with human conversational norms (Q. Huang et al., 2024).

Large model-based AI arguably demonstrates the highest level of agency observed to date, as they fulfill all four core conditions more comprehensively than other paradigms. These models interact dynamically with users, generating responses based on prompts and contextual cues. They exhibit autonomy by producing outputs without step-by-step human guidance and show adaptability by refining outputs through user feedback and continuous training. They also demonstrate goal-directed behavior by optimizing responses to meet defined objectives (Y. Wang et al., 2024). Although users initially define these goals, the models increasingly display emergent properties, structuring responses in ways that suggest long-term reasoning (Acharya et al., 2025). These systems can also cascade planning, where high-level objectives (e.g., managing a meeting) are decomposed into sub-tasks (e.g., scheduling participants, setting an agenda, transcribing, summarizing discussions, generating action items, and following up). This hierarchical task structure allows them to approximate goal-directed behavior more closely, as they execute sequences of actions independently rather than relying on human prompts for each step. Known as chaining, this ability to break down and execute complex tasks sequentially marks a key aspect of their agentic capabilities (X. Huang et al., 2024). Additionally, some models employ "chain-of-thought" reasoning, generating intermediate reflective steps before producing final outputs – enhancing their performance on multi-step problems and making their reasoning more interpretable to humans (Cheng et al., 2024).

In summary, AI systems already exhibit agentic behaviors without intrinsic intentionality. This viewpoint evaluates agency based on functional and observable behaviors rather than mental representations, enabling an assessment of AI's capabilities independent of human-like cognition. AI systems increasingly demonstrate agency by executing complex tasks without continuous human intervention, adapting their outputs to changing inputs, engaging in dynamic interactions with users and environments, and optimizing toward pre-defined objectives. While these functions remain mostly externally imposed, ongoing advancements suggest that the agenticness of AI systems will intensify as their capabilities expand.

# Typological Framework Building Approach

The phases of our approach are outlined in Table 1. Aligned with our objective, we develop a theoretical typology to integrate existing knowledge and anticipate future developments of agentic AI. Typologies systematically categorize theoretical dimensions into distinct types, supporting further theorization by establishing meaningful relationships (Bailey, 1994). Given the evolving complexity of agentic AI, this approach abstracts key characteristics while remaining adaptable to emerging trends. It bridges theoretical constructs with empirical observations, offering a foundation for ongoing research (Delbridge & Fiss, 2013; Negoita et al., 2018). Unlike taxonomies, which classify existing entities, typologies provide a forward-looking lens for conceptualizing and theorizing phenomena that have yet to materialize (Bailey, 1994).

| Table 1. Summary of the typology building approach ||
|---|---|
| **Methodology** | Construction of a conceptual typology (Bailey, 1994) |
| **Purpose** | Characterize and enable the exploration of agentic AI systems |
| **Theoretical foundation** | • Minimal agency (Barandiaran et al., 2009)<br>• Levels of abstraction of "agenthood" in artificial agents (Floridi & Sanders, 2004) |
| **Construction quality criteria** | The following criteria guide the construction process (Mills & Scott, 1983):<br>(a) *Intuitively sensible*: It should capture the common sense of an agentic AI system by grouping systems that exhibit similar levels of agency while separating those that differ. Also, the names of the dimensions and characteristics should be mostly self-explanatory to readers who are up to date on recent AI developments.<br>(b) *Collectively exhaustive*: Should accommodate the classification of all AI systems – including those with minimal or nonexistent agency – ensuring that their inherent differences become apparent.<br>(c) *Mutually exclusive*: The typology offers a set of mutually exclusive characteristics, ensuring unambiguity and internal consistency, with no overlap across characteristics within a dimension.<br>(d) *Construct validity*: The typology should demonstrate discriminant validity – that is, it should be sufficiently distinct from related typologies, which we confirm in phase 1.<br>(e) *Conceptually elegant*: Although conceptual elegance is partly subjective, it involves minimizing the number of dimensions and characteristics in a typology, enhancing comprehension. |
| **Phase 1**<br>**Establish construct validity** | • Review related work to position the study within existing typologies and classifications.<br>• Establish the typology's construct validity by identifying conceptual overlaps and gaps, thereby defining the specific contribution and focus of the proposed typology. |
| **Phase 2**<br>**Construction of the ideal type** | • Concept-matrix-based approach to extract key dimensions of agentic AI functionalities (Webster & Watson, 2002)<br>• Extrapolation of ideal forms as extremes of the dimensions, forming the ideal type (Bailey, 1994) |
| **Phase 3**<br>**Substruction** | • Assigning characteristics to dimensions extracted from the concept matrix<br>• Ensuring mutual exclusivity across the dimensions<br>• Merging and adjustment of dimensions |
| **Phase 4**<br>**Refinement** | • Refine the typology to ensure conceptual clarity, balance, and adherence to quality criteria.<br>• Introducing ordinal progression with levels from non-agentic to more advanced levels<br>• Terminological adjustments and restructuring to enhance self-explanatory consistency and ensure comprehensive coverage of agentic AI systems |
| **Phase 5**<br>**Evaluation and refinement with empirical types** | • Evaluation of typology with real-world objects (Szopinski et al., 2019)<br>• Application of a human-AI hybrid approach<br>• Employing OpenAI Deep Research to validate typology with 43 real-world AI systems |
| **Phase 6**<br>**Reduction into constructed types** | • Exemplary simplification and reduction of the final typology into constructed types<br>• Demonstration that the typology can be adapted to different purposes |

To build a robust typological framework for characterizing and exploring agentic AI systems – both existing and emerging – we first establish a theoretical foundation grounded in the conditions of artificial agency discussed earlier. The typology-building process is primarily informed by Bailey (1994), with methodological extensions (Göldi & Rietsche, 2024; Webster & Watson, 2002). We apply a multi-phase approach, beginning with constructing an ideal type based on existing literature. The ideal type functions as a conceptual benchmark that guides the refinement and expansion of the typology in later phases. Table 1 outlines the criteria used for constructing the typology (Mills & Scott, 1983).

### *Phase 1: Establish Typoloy's Construct Validity with Related Work*

In the first phase of our construction process, we aim to establish the (d) construct validity of the proposed typology by ensuring it is distinguishable from existing typologies and related classification frameworks (Mills & Scott, 1983). To achieve this, we conducted a literature review (Webster & Watson, 2002) using three scientific databases – *Web of Science*, *Scopus*, and *AISeL* – with search terms including "AI", "artificial intelligence", "large model", and "large language model", each combined with "agentic" and "agent". We extended our search to *arXiv*, a widely used computer science and engineering repository for research papers and pre-prints to capture the most recent developments in agentic AI. Given the fast-paced nature of the field, including *arXiv* is particularly relevant to our research focus on emerging technologies and aligns with IS literature review practices that increasingly incorporate sources from fast-moving publication venues (e.g., Albrecht et al., 2023; Antunes & Tate, 2024). Following established guidelines for including non-peer-reviewed literature in qualitative research, we evaluated each arXiv source for author authority, objectivity, and methodological soundness (Garousi et al., 2019).

We prioritized recent overview papers – published in 2023 or later only to include the most recent advancements in AI – such as classifications, categorizations, surveys, reviews, typologies, or similar constructs synthesizing characteristics of agentic AI systems. This approach was aligned with the purpose of constructing a forward-looking ideal type rather than conducting an exhaustive backward-looking systematic literature review. From the results, we identified twelve particularly relevant papers: **1** = (L. Wang et al., 2024); **2** = (Durante et al., 2024); **3** = (Zhao et al., 2023); **4** = (Acharya et al., 2025); **5** = (Händler, 2023); **6** = (Cheng et al., 2024); **7** = (Y. Wang et al., 2024); **8** = (Xie et al., 2024); 9 = (Göldi & Rietsche, 2024); **10** = (Q. Huang et al., 2024); **11** = (Xi et al., 2025); **12** = (X. Huang et al., 2024).

In summary, the papers map the fast-moving agentic-AI landscape along markedly different axes, such as an emphasis on autonomy, multimodality, goal complexity, architecture, or application scenarios. L. Wang et al. (2024) survey LLM agents, emphasizing architecture and application scenarios. The paper offers a state-of-the-art overview of how these agents are constructed and function. Zhao et al. (2023) follow a similar approach but with a narrower scope, focusing more deeply on the core components of LLM agents. Similarly, X. Huang et al. (2024) narrow their survey to planning mechanisms. Cheng et al. (2024) extend this by offering precise definitions, a taxonomic perspective, and application scenarios. Xi et al. (2025) also detail agent construction and uniquely expand the discussion to include practical deployments and a future vision of "agent societies." While these works align partially with our effort to map the landscape of agentic AI, they remain limited by their exclusive focus on LLM-based agents. For example, Göldi and Rietsche (2024) propose a typology of LLM-based agents, enabling comparison across varying sophistication levels and projecting future trajectories. However, like others, it does not facilitate comparisons between agentic and non-agentic systems and lacks an agnostic lens for different AI paradigms.

Beyond LLM-centric studies, Durante et al. (2024) offer a multimodal and holistic overview by incorporating various perspectives, such as external knowledge, multi-sensory inputs, and human feedback, yielding a holistic view of agentic AI, across paradigms, learning strategies, different agent types, and applications. Xie et al. (2024) adopt a taxonomic approach focusing on multimodal agents, their core components and methodologies, and their assessment and evaluation. Q. Huang et al. (2024) introduce the "holistic intelligence" concept driven by agentic AI and propose a cross-domain framework grounded in interactive manipulation and embodied operation. Other studies concentrate on operational aspects. Acharya et al. (2025) survey various topics such as agentic AI structure, applications, evaluation, concerns, and responsible implementation. Händler (2023) adds to the understanding of multi-agent systems by examining the balance between autonomy and alignment – a critical factor in agent design. Y. Wang et al. (2024) comprehensively survey the security and privacy concerns in connected agentic AI systems, detailing architectures, threats, countermeasures, and open research challenges.

While these contributions illustrate the richness, dynamics, and complexity of agentic AI systems, they fall short of grounding their agency in a coherent theoretical framework. Much of the existing research consists of technical overview papers that focus on system architectures, components, or application domains, but they do not offer a model- or technology-agnostic blueprint for understanding agentic AI. In contrast, our work anchors its typology in the functional dimensions of agency – interactivity, autonomy, adaptability, and normativity – whereas previous studies treat agentic behavior as a system-level capability rather than a principled concept. Although each study contributes to the broader understanding of agentic AI, most do not place agency itself at the center of their synthesis. Our typology aims to address this gap by applying canonical agency conditions to position AI systems along a continuum from passive tools to fully agentic entities, regardless of the underlying model. This theory-driven approach brings construct validity (d) to categorizing agentic AI and equips researchers and practitioners with a more systematic way to interpret exisiting and anticipate developments in this emerging field.

## *Phase 2: Construction of the Ideal Type*

Defining an ideal type is a well-established first step in building typologies grounded in theoretical or conceptual foundations. Originating from Weber (1949), ideal types represent the purest form of a type's characteristics – not as utopian ideals but as extreme cases that fully express each dimension. While empirical types (i.e., real-world cases) may exist in some contexts, an ideal type may never fully manifest in reality. They provide a valuable conceptual tool for guiding theoretical exploration. Their primary purpose is to serve as a starting point in the typology construction process (Bailey, 1994).

| Table 2. Concept matrix | | | | | | | | | | | | |
|---|---|---|---|---|---|---|---|---|---|---|---|---|
| **Characteristics of agentic AI systems** | 1 | 2 | 3 | 4 | 5 | 6 | 7 | 8 | 9 | 10 | 11 | 12 |
| Action impact (changing environment, altering internal state, triggering new action) | x | | | | | | | | | | | |
| Action space (only output/internal, external tools, create new tools, other agents, physical environments, virtual environments) | x | x | | x | x | x | x | x | | x | x | |
| Action target (task, exploration, communication) | x | | | | x | | | | | | | |
| Autonomy to operate (static, adaptive, self-organizing, short-term, checkpointing, continuous operation) | x | | x | x | x | | | | x | | | |
| Capability expansion (trial-and-error, crowd-sourcing, experience, self-driven) | x | | | | | | | | | | | |
| Embodied actions (observation, manipulation, navigation) | | x | | | | x | | | | x | x | |
| Emotional understanding and empathy | | x | | | | | | | | | | |
| Fine-tuning (human, generated, real-world data sets) | x | x | | x | | | | | | | | |
| Improvement or adaptability (human feedback, other models, orchestration modification, self-update, adaptive-control mechanisms) | | x | | x | | | | | x | x | | |
| Interaction module (agent-agent, agent-human, agent-environment) | | x | | | x | x | x | | | | | |
| Interactivity level (only input/output, during processing, proactive engagement) | | | | | | | | | x | | | |
| Knowledge scope (commonsense, domain, retrieval-augmented generation, dynamic narrowing, continuously updating, active probing) | | x | | x | | x | | | x | | x | |
| Memory operation (reading, writing, reflection, retrieval) | x | x | | x | | x | | | | | x | |
| Memory structure (Input, Training memory, Short-term memory, long-term memory, hybrid memory) | x | x | x | x | x | x | x | x | | | x | x |
| Memory format (languages, databases, lists, embeddings) | | x | | | | x | | | | | x | |
| Multi-agent collaboration or infrastructure | | x | | x | x | x | x | x | | | | |
| Perception or modality support (text, visual, auditory, other futuristic such as touch or smell, multimodality) | x | x | x | x | x | x | x | x | x | x | x | |
| Personalization (None, manual, memory-based, profile-based) | x | | | | x | | | | x | | | |
| Planning formulation or reasoning (In context-learning or single-path, multi-path, via external tool, task decomposition) | x | x | x | x | x | x | x | x | x | x | x | x |
| Planning reflection, rethinking, feedback, inspection, or introspection (environment, human, self-reflection) | x | | x | | x | x | | | | x | x | x |
| Profile attainment (crafted by humans, based on empirical data, self-generation) | x | | | | x | | | | x | | | |

To construct the ideal type, we extracted key characteristics of agentic AI systems from the literature (Table 2), drawing on the results of Phase 1 and following established approaches (Göldi & Rietsche, 2024; Negoita et al., 2018). We applied a concept matrix-based method (Webster & Watson, 2002) to identify attributes specifically aligned with the four agency conditions that form our theoretical foundation. These characteristics were then synthesized into common themes representing core functionalities and dimensions of agentic AI systems (italics for readability). This approach also supports criterion (e) of conceptual elegance by reducing the number of concepts while maintaining theoretical clarity.

A key theme in the literature is that agentic AI systems exhibit *reasoning* capabilities through multi-step planning. This ability enables such systems to decompose complex tasks into smaller sub-tasks, allowing for structured, long-term execution (L. Wang et al., 2024). The planning process often includes *self-evaluation* – referred to as rethinking, introspection, or reflection – where the system assesses and adjusts its sub-tasks based on human input or environmental feedback to improve the accuracy and quality of outcomes (Zhao et al., 2023). Planning is supported by internal *knowledge* acquired during training, such as domain-specific expertise or commonsense reasoning, and may also be enhanced through external knowledge sources like retrieval-augmented generation (Göldi & Rietsche, 2024). In addition, *perception* capabilities are central to agentic AI systems, which may rely on a single modality (e.g., visual input) or integrate multiple modalities, including language, spatial awareness, audio, and emotional signals – enabling more human-like cognition (Zhao et al., 2023).The literature consistently identifies *memory* as a fundamental capability of agentic AI systems. Agentic AI often incorporates short-term memory for tracking planning steps and long-term memory to reflect on previous interactions and adapt future behavior accordingly (Acharya et al., 2025). Rather than passively processing input and output, agentic AI systems can take *action* upon their environment based on the outcomes of their internal planning processes (L. Wang et al., 2024). These systems also demonstrate *interactive* behavior, such as engaging proactively with human users, collaborating with other AI agents, and sometimes interfacing with physical systems. Another essential characteristic is *contextual awareness*, where agentic AI adjusts its behavior depending on situational context – for example, personalizing interactions with individual users or modifying responses based on assigned roles, sometimes resembling shifts in personality (Göldi & Rietsche, 2024). *Self-improvement* is also a defining feature, allowing agentic AI to learn from experience and refine its performance over time (Acharya et al., 2025). Finally, *autonomy* enables these systems to operate independently and, in many cases, continuously within their functional domain without human oversight (Acharya et al., 2025).

These characteristics were refined into idealized dimensions to define the ideal type, which serves as a conceptual reference point for developing the typology, as summarized in Table 3. This ideal type is neither exhaustive nor final; following the approach of Göldi and Rietsche (2024), it functions as a foundation for further refinement in subsequent phases.

| Table 3. The ideal type of agentic AI systems | |
|---|---|
| **Dimension** | **Idealized characteristic** |
| Knowledge | Dynamically actively updates its knowledge base by integrating domain-specific expertise, general commonsense reasoning, and real-time external information from databases, APIs, and other sources to enhance decision-making. It can explore the knowledge space to probe for new, relevant information. |
| Perception | Processes multimodal inputs, including text, images, audio, and spatial data, to build an adaptive understanding of its environment. Can detect human emotions, sentiment, and non-verbal cues to refine interaction quality. |
| Reasoning | Decomposes complex goals into structured multi-step planning strategies and anticipates future scenarios to ensure efficient long-term decision-making. |
| Self-evaluation | Continuously assesses its reasoning, refining planning strategies through introspection, iterative adjustments, and feedback from interactions with its environment, other agents, or humans. |
| Action space | Executes embodied tasks autonomously across digital, physical, and multi-agent environments. |
| Autonomy | It operates entirely as an individual-like entity independent of human or other external intervention. |
| Interactivity | Engages proactively and continuously with humans, other AI agents, and systems. It is nested as an actor in a social system. |
| Memory | Maintains sophisticated memory structures, such as short-term memory for task execution and long-term memory for learning, storing, and retrieving relevant information. It also reflects on past experiences. |
| Contextual awareness | Adjusts behavior and communication style dynamically and intuitively based on context to align with task requirements and user expectations. |
| Self-improvement | Learning from feedback, self-assessment, and new data, refining decision-making processes, optimizing performance, adapting its operational capabilities over time, and rewriting its code-base or evolving new model architectures. |

## *Phase 2: Substruction into Dimensions and Characteristics*

Substruction focuses on consolidation, resolving contradictions, and terminological precision. While these activities aim to produce a coherent set of dimensions to work toward fulfilling (a) intuitive sensibility and (e) conceptual elegance, the core of substruction lies in systematically varying the presence, absence, or degree of characteristics across the dimensions (Bailey, 1994). It ensures that every dimension is discriminatory (i.e., reveals variations) and comprehensive (i.e., covers the range variation) while adhering to (c) mutual exclusivity (Mills & Scott, 1983). To achieve this, we followed an iterative approach similar to what is outlined by Nickerson et al. (2013). However, instead of striving for an exhaustive overview that categorizes every technical aspect of empirical agentic AI systems (as one might do in a technical synthesis or taxonomy), we aimed to produce a typology to explore agenticness at its core.

We mapped the concepts extracted from the literature (Table 2) to the dimensions defined in the ideal type, aiming to meet the criteria of (c) mutual exclusivity and (b) collective exhaustiveness. During this process, we encountered challenges such as overlapping characteristics and inconsistent levels of measurement – some features coincided, while others reflected ordinal progression. These issues prompted adjustments to the dimensional structure. We merged *reasoning* and *self-evaluation*, as the latter could not be isolated into standalone characteristics but instead saw it manifest as technical methods that were neither universally applicable nor self-explanatory. Self-evaluation was therefore integrated as a characteristic within the reasoning dimension, distinguished from task decomposition, which lacks internal reflection. We also removed *memory* as a dimension since short- and long-term memory and input embeddings are not mutually exclusive (Zhao et al., 2023); instead, memory-related traits were embedded within contextual awareness and self-improvement. Finally, we merged *action space* with *interactivity* due to their substantial conceptual overlap, consolidating how agentic AI systems engage with their environment.

## Phase 3: Refinement with ordinal levels

After establishing an initial typology of agentic AI systems, we aimed to refine the typology to satisfy all five criteria sufficiently. While reviewing the current configuration, we identified three main areas for improvement. First, we noticed a significant imbalance in the number of characteristics across the dimensions, signifying that it was not (a) conceptually elegant. To address this, we decided to balance all dimensions to ensure more effective separation between different agentic AI systems. Second, the typology only characterized agentic properties in their current form, preventing us from comparing partially or non-agentic systems. This shortcoming meant it did not adhere to (b) collective exhaustive. Third, we observed that some of the characteristics were nominal and descriptive, which did not adequately capture similar levels across dimensions, nor did it enable us to account for future trajectories.

To address this, we first introduced a new characteristic as a baseline, "non-agentic," in each dimension, creating an ordinal progression that distinguishes "basic" and more "sophisticated" agentic capabilities. Finally, to acknowledge potential future developments, we added a fourth ordinal level, "general intelligence," rooted in the concept of artificial general intelligence (AGI) as a speculative evolution across dimensions. To define appropriate characteristics at this level, we referred to human or sentient-like traits and extended these with insights from AGI research (Mitchell, 2024). As part of these enhancement efforts, we also revised the terminology for dimensions and characteristics to ensure they are self-explanatory. At this point, we revisited the five outlined criteria in Table 1 within the author team. We decided that the restructured typology (refer to Table 5 for the final typology after evaluation) provides a comprehensive framework for understanding agentic AI, leading us to the next phase of evaluating with empirical types.

## Phase 4: Evaluation and Refinement with Empirical Types

A proven method for assessing conceptual frameworks' practical applicability and validity involves applying and mapping real-world objectives against their dimensions and characteristics (Szopinski et al., 2019). We decided to use a human-AI hybrid approach, which can improve the quality output of tasks (Fügener et al., 2022). Specifically, we focused on identifying empirical types of agentic AI systems, extracting detailed descriptions, and mapping them to the current typology – tasks particularly well-suited for research-focused AI systems. Recent research has demonstrated that while current AI systems are not yet competent to perform fully automated qualitative analysis, they can function effectively as screening, advisory, or coding tools while keeping humans in the loop. For example, McCarthy et al. (2024) demonstrate how an agentic AI system can support grounded theory-based research. We relied on *Humanity's Last Exam* and *GAIA* benchmarks to select a suitable AI system to support Phase 4, which both evaluate AI agents on expert-level tasks. OpenAI's Deep Research (DR) ranks highly on these benchmarks and is designed explicitly for synthesizing large amounts of information and completing multi-step research tasks while citing each claim. It is commercially available through OpenAI's pro subscription (OpenAI, 2025b).

We conducted multiple test runs to fine-tune DR's reasoning process, testing different prompt designs to balance explicit instruction with flexibility. Based on these iterations, we developed a four-step reasoning strategy within a single prompt (Table 4). First, DR was instructed to build a dataset of empirical agentic AI systems, including widely known and research-stage models. Each system must satisfy at least one of the four agency conditions defined. DR was also required to justify each inclusion using detailed descriptions from peer-reviewed articles or technical research papers, explicitly linking its reasoning to the agency conditions. Secondly, DR mapped the typology developed in Phase 3 to each empirical type and explained the rationale for each classification. The third step asked DR to provide feedback on the typology based on this mapping experience. Finally, DR was instructed to propose a revised version of the typology, drawing on its previous steps. To ensure consistency and robustness, we had DR repeat the whole process thrice, reflect on its prior outputs, incorporate additional empirical types, and revise the results if needed.

In summary, DR introduced several meaningful changes to the typology, but the overall structure remained stable—indicating that our conceptual typology from Phase 3 was already well-founded. The most notable addition was the dimension of *normative alignment*, which addresses how and to what extent an agentic AI system manages goals, ethical considerations, and operational constraints in its reasoning or behavior. Interestingly, this dimension had already emerged in one of our earlier test runs, further supporting the validity and transferability of the results. Other changes primarily focused on refining the nomenclature of dimensions and characteristics. During the two reflection cycles, no structural modifications were made, apart from adjustments to terminology and refinements to the ordinal levels, to enhance clarity and conceptual elegance. DR evaluated 43 agentic AI systems as part of its reasoning process. Details on the prompts used and the systems are provided in the supplementary material.

| Table 4. Summary of Human-AI hybrid approach for phase 4 | | |
|---|---|---|
| **Steps** | **OpenAI Deep Research tasks** | **Author tasks** |
| Dataset | • Search for both well-known and research-stage AI systems.<br>• Must fulfill at least one of the conditions of agenticness as outlined in the theoretical foundation section<br>• The decision to include a system in the data set should be based on detailed information, such as in peer-reviewed publications or research papers.<br>• Explain the decision criteria of why an AI system was included | Verification of results by checking the references and decision criteria as outlined by DR |
| Mapping | • Map the dataset of agentic AI systems against the current typology.<br>• Outline and explain the reasoning for mapping. | Checkpointing for inconsistencies, hallucinations, or logical errors, as well as reviewing the reasoning given by DR for the performed steps |
| Feedback | • Give feedback on the current typology<br>• Outline and explain the reasoning for the feedback | |
| Suggestion | • Suggest a newly revised typology<br>• Reflect on the construction criteria | |
| Three reflection cycles | • Repeat the reasoning steps<br>• Compare the new results to the previous version and explain the reasoning for adjusting the typology | |

## *Phase 5: Reduction into Constructed Types*

We performed one reduction phase as the final phase. The objective was to reduce the complexity of the typology without removing the core aspects or invalidating its complete version (Bailey, 1994). A reduced typology can serve a specific purpose, such as adapting it for practitioners (e.g., Göldi & Rietsche, 2024). Thus, the result of the reduction phase is not exhaustive or final but rather an illustration of how one might adjust the typology or use it for a specific context. For this, we applied the concept of constructed types, which are not an extreme or accentuated form but rather a common abstraction of an empirical type (Bailey, 1994), which could also be referred to as something akin to an "archetype". We reflected upon the eight dimensions for common themes or relations between them. One critical insight was that some dimensions focus on an AI system's internal properties while others reflect how the AI operates externally in its environment. Building on that observation, we recognized a grouping into two overarching dimensions:

- *Cognitive agency*: Encompasses the dimensions of reasoning, knowledge scope, self-improvement, and normative alignment, describing how the AI system "thinks" – the extent of its internal deliberation, capacity to update or refine its strategies, and the degree to which it considers norms in pursuit of its goals.
- *Environmental agency*: includes the dimensions of perception, operation, interactivity, and contextualization, describing how the AI system acts in or interprets its surroundings – its ability to sense external inputs, engage with humans or other agents, how it is nested and operates in its environment, and place external information into a meaningful context.

Through this reduction, we arrive at a two-axis framework that, rather than replacing the original dimensions, provides a streamlined lens for classifying AI systems by their key capabilities (Figure 1).

*Exploring Agentic AI Systems*| | |
|---|---|
| **Table 5. Final typology after evaluation with empirical types** | |
| **Dimension** | **Short description** |
| **Knowledge scope** | **The extent and domain of information an AI system has access to** |
| Narrow (0) | The AI system's knowledge is limited to a specific domain or task, with fixed information developers provide (no capability to go beyond its training data or pre-programmed knowledge). |
| General (1) | The AI system possesses broad, pre-trained knowledge covering multiple domains but does not actively fetch new external data (i.e., knowledge is fixed at deployment). |
| Externally-informed (2) | The AI system can dynamically incorporate new information from external sources during operation. For example, it can query databases, read documents or sensor inputs (retrieval-augmented generation), and use that information to update its reasoning process. |
| Exploratory (3) | The AI system actively seeks out or discovers new knowledge through its actions. It can conduct experiments, explore environments, or ask questions to expand what it knows. This level implies open-ended learning beyond the information given. |
| **Perception** | **The AI's ability to perceive inputs** |
| None (0) | The AI system has no perceptual input from its environment; it only processes a confined input, e.g., input variables and labels. |
| Unimodal (1) | The AI system perceives one type of input (e.g., purely language, purely vision) and relies on this single stream for operation. |
| Multimodal (2) | The AI system can handle multiple types of inputs and combine them (including transcription of other modalities, such as in some LLMs). |
| Intuitive (3) | The AI system demonstrates human-like intuitive perception, grasping abstract or high-level features. It holistically fuses multiple input modalities into a coherent understanding of its environment. |
| **Reasoning** | **The AI's capability to process and plan tasks** |
| One-shot (0) | The AI system maps each input to an immediate output. It produces one-shot responses based on learned patterns, rules, or weights. |
| Decompositional (1) | The AI system plans tasks by decomposing them into subgoals or by searching through possible action sequences. It does not reflect on its reasoning but does purposeful, stepwise problem-solving. |
| Reflective (2) | The agent has some meta-cognitive loop: it can examine interim results, revise its strategies, and potentially correct mistakes mid-process (e.g., iteratively refining plans or using chain-of-thought reasoning). |
| Theory-of-Mind (3) | The AI models other agents' mental states – their beliefs, desires, or intentions – and plan accordingly. This involves social/strategic reasoning, cooperation, or negotiation. |
| **Interactivity** | **The degree to which and how an AI system engages with its environment** |
| Passive (0) | The AI system does not take action in the environment. |
| Tool-using (1) | The AI can invoke or use external tools or APIs to achieve tasks but relies on pre-defined resources. |
| Tool-building (2) | The AI system generates new tools or methods (e.g., writing code or macros). |
| Dynamic (3) | The agent freely engages with a complex, open-ended environment or multiple actors, such as collaborative interactions, competitive engagements, or social dialogues. |
| **Operation** | **The operational mode in which the AI system functions** |
| On-demand (0) | The AI system only acts when explicitly invoked (e.g., a function that returns a result upon request). It remains idle otherwise. |
| Periodic (1) | The AI system operates in discrete episodes or sessions (e.g., on a per-game or per-conversation basis) and is inactive outside those time frames. |
| Continuous (2) | Once activated, the AI system runs uninterrupted in real-time within its specified goal-setting, processing inputs and issuing outputs as needed until it is stopped. |
| Self-organizing (3) | The AI system not only runs continuously, but it self-manages its activities and goals over time. It decides when and what to do without external triggers by spawning new sub-agents, reallocating roles, or modifying its operational architecture. |
| **Contextualization** | **The AI's ability to integrate and retain context** |
| Stateless (0) | The AI system handles each task or input in isolation, with no awareness of prior context or changes in the environment. |
| Local (1) | The AI system considers the immediate environment or short-term context (e.g., the conversation and sensor reading). It does not store or recall older information. |
| Memory-based (2) | The AI system stores and retrieves relevant information about past states or events to guide current decisions (e.g., conversation history, skill library, short- or long-term memory). |
| Holistic (3) | The AI system exhibits deep, integrated context comprehension akin to human-like situational awareness. It can rapidly grasp the meaning of novel scenarios. |
| **Self-improvement** | **The AI's ability to learn and adapt** |
| Static (0) | The AI system's policy or parameters never change once deployed; no learning occurs during operation. |
| Reactive (1) | The AI system makes minor adjustments (e.g., switching among pre-defined modes) based on immediate feedback but does not truly acquire new knowledge or update its long-term model. |
| Adaptive (2) | The AI system updates its capabilities over time (e.g., online learning, storing and leveraging new information). It improves or changes its behavior through experience without requiring external re-training. |
| Evolutionary (3) | The AI system restructures or evolves more profoundly, possibly rewriting its operational procedures or generating new "descendant" models. It can recursively self-improve in an open-ended process. |
| **Normative alignment** | **The AI's alignment with ethical, social, or procedural norms** |
| Unaware (0) | The AI system operates solely according to its functional goals without explicitly considering norms. |
| Rule-bound (1) | The AI system follows explicit rules (e.g., policies) if they are clearly stated or hard-coded. It does not understand the rationale behind them and cannot handle novel conflicts in norms. |
| Norm-aware (2) | The AI system can understand and apply social or ethical norms in context, adjusting behavior appropriately. It recognizes typical expectations and constraints. |
| Value-aligned (3) | The AI system proactively aligns decisions with broader ethical principles or human values, even in novel or conflicting situations. It can reason about moral trade-offs akin to a well-informed human's principled judgment. |
| 0 = non-agentic; 1 = basic; 2 = sophisticated; 3 = general intelligence or "AGI-like" (speculative) | |

*Pacific-Asia Conference on Information Systems, Kuala Lumpur 2025*
11

## Applying the Typology

Following the six phases, we present the typology of agentic AI systems scoped according to our definition of agenticness. While it represents a complete iteration, it is not final. As with all typological research, it can be revised to reflect future developments or shifts in relevance. It consists of eight dimensions, each with four characteristics representing a progression from non-agentic to agentic AI: (0) non-agentic, (1) basic, (2) sophisticated, and (3) general intelligence (futuristic). A detailed overview is provided in Table 5.

| Table 6. Classification of Empirical Types | | | |
|---|---|---|---|
| System | Cognitive Agency | Environmental agency | Constructed Type |
| **Deep Research** | Knowledge Scope: **externally-informed (2)**<br>Reasoning: **reflective (2)**<br>Self-Improvement: **reactive (1)**<br>Normative Alignment: **norm-aware (2)** | Perception: **unimodal (1)**<br>Interactivity: **passive (0)**<br>Operation: **periodic (1)**<br>Contextualization: **local (1)** | **Research agent** |
| **Copilot Agents** | Knowledge Scope: **externally-informed (2)**<br>Reasoning: **one-shot (0)**<br>Self-Improvement: **reactive (1)**<br>Normative Alignment: **norm-aware (2)** | Perception: **unimodal (1)**<br>Interactivity: **tool-using (1)**<br>Operation: **continuous (2)**<br>Contextualization: **memory-based (2)** | **Task agent** |
| **Copilot Chat** | Knowledge Scope: **externally-informed (2)**<br>Reasoning: **One-shot (0)**<br>Self-Improvement: **reactive (1)**<br>Normative Alignment: **rule-bound (1)** | Perception: **unimodal (1)**<br>Interactivity: **passive (0)**<br>Operation: **on-demand (0)**<br>Contextualization: **local (1)** | **Simple agent** |
| **Operator** | Knowledge Scope: **externally-informed (2)**<br>Reasoning: **reflective (2)**<br>Self-Improvement: **adaptive (2)**<br>Normative Alignment: **norm-aware (2)** | Perception: **multimodal (2)**<br>Interactivity: **tool-using (1)**<br>Operation: **continuous (2)**<br>Contextualization: **local (1)** | **Complex agent** |

To illustrate its real-world applicability, we applied the typology to classify four agentic AI systems (Table 6). As these systems are under active development, the classification reflects their state as of February 2025. We selected systems not included in Phase 4, using a hands-on approach that combined direct testing with an analysis of technical documentation from developers. Some dimensions were challenging to assess due to limited public information or proprietary constraints – such as self-improvement or normative alignment potentially being embedded in the model but not fully disclosed. Where necessary, we provided informed estimations based on testing experience to demonstrate the typology's practical use.

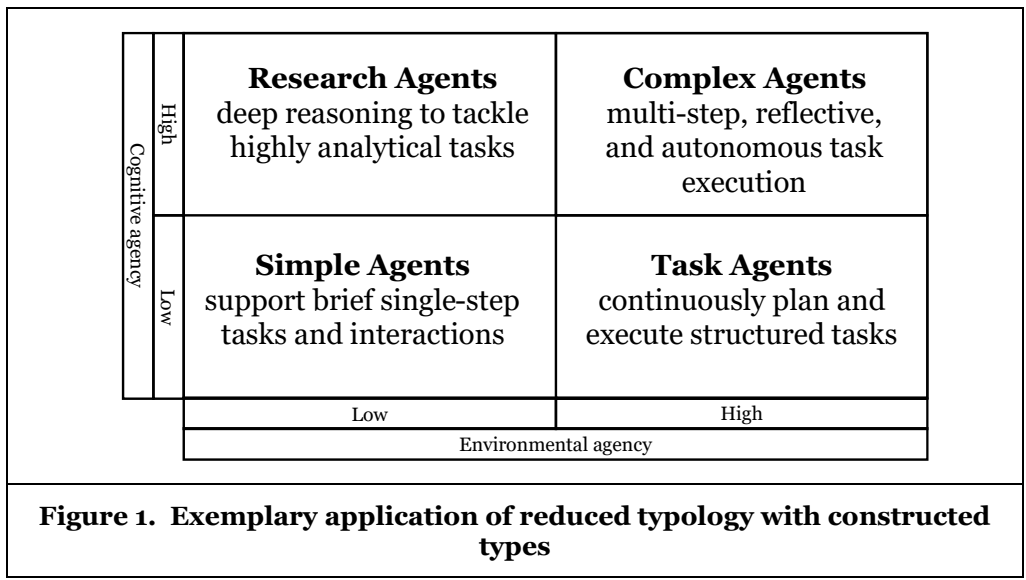

**Figure 1. Exemplary application of reduced typology with constructed types**

**Deep Research.** DR is based on the o3 model — a reflective GPT successor to o1 — using chain-of-thought reasoning to perform multi-step analyses (Reasoning: 2). It can access external web sources and user-uploaded PDFs (Knowledge Scope: 2) but remains unimodal, processing only text (Perception: 1). DR cannot act in the environment and is limited to retrieving and synthesizing content upon user request (Interactivity: 0). It functions in discrete sessions that reset after each task (Operation: 1), retaining local context only within a session (Contextualization: 1). The system adapts reactively to user feedback but is only updated offline by developers (Self-Improvement: 1). DR is norm-aware (Norm Alignment: 2), guided by safety and ethical standards embedded during its development (OpenAI, 2025d).

**Copilot Agents.** Copilot agents are an extension of the M365 Copilot, enabling users to automate specific tasks with agentic AI capabilities. They handle clearly defined and repeatable, one-shot tasks (Reasoning: 0) and can incorporate internal or external data as defined at the creation of the agent (Knowledge Scope: 2). However, their perception remains unimodal by focusing on textual content (Perception: 1). Copilot agents enable the user to integrate a large number of pre-defined tools but can also access custom connectors to take actions in various environments (Interactivity: 1). They can operate continuously (Operation: 2), persisting across various triggers or events until a human deactives it. They maintain memory-based context (Contextualization: 2) across sessions, retaining relevant details. These agents adapt reactively when a user adapts the settings of the agent (self-improvement: 1). Finally, they are norm-aware (norm alignment: 2), as they reflect organizational governance settings (e.g., data retention, sensitivity labels) and built-in security rules to ensure compliance and acceptable usage (Microsoft, 2025b).

**Copilot Chat.** Copilot Chat provides one-shot Q&A without advanced reasoning (Reasoning: 0). It uses an LLM with broad knowledge and has access to the web (Knowledge Scope: 2), focuses on text (Perception: 1), remaining passive (Interactivity: 0) and operating on-demand (Operation: 0). It retains local context (Contextualization: 1) during each session but does not preserve information across sessions. Its self-improvement is reactive (1), as users can give immediate feedback to reiterate a task with additional information, and it is governed by pre-defined rules (norm alignment: 1) (Microsoft, 2025a).

**Operator.** Operator is built on a vision-augmented GPT-4 called Computer-using Agent (CUA) that uses stepwise, reflective reasoning for multi-step tasks on the web (Reasoning: 2). It references live content from the web (Knowledge Scope: 2), classifying it as externally informed and processes both text and screenshots (Perception: 2). Operator navigates and fills online forms in a tool-using capacity (Interactivity: 1), remaining active continuously (Operation: 2) until stopped by the user. It retains local context (Contextualization: 1) within each session – recalling webpage states and user instructions – but does not store data across separate runs. Its self-improvement is adaptive (2) by self-correcting mistakes or human guidance during a session, while model-wide updates rely on offline developer interventions. Finally, Operator is norm-aware (Norm Alignment: 2) as it has various inherent safeguards, such as human takeover mode and safety mechanisms against malicious websites or other threats (OpenAI, 2025c).

We applied the reduced typology to demonstrate how it can be used alongside empirical types to deduce exemplary constructed types (Figure 1). Four types emerge based on whether a system shows low or high cognitive and environmental agency, using a threshold of 6 points to indicate high agency. *Simple agents* (e.g., Copilot Chat) display low or no agency across most dimensions typically focused on basic knowledge retrieval or one-off responses. *Research agents* show high cognitive agency through reasoning and reflection but cannot act in the environment (e.g., supporting complex tasks for knowledge workers without automation). *Task agents* (e.g., Copilot Agents) operate continuously using pre-defined data sources, triggers, and tools, focusing on executing structured, repeatable tasks rather than complex reasoning. *Complex agents* (e.g., Operator) begin with a broad user-defined objective and autonomously initiate reflective reasoning, searches, or tool use — actions not explicitly defined by the user, unlike task agents.

In summary, we have demonstrated how our typology's complete eight-dimensional model and reduced two-axis perspective can be applied to real-world agentic AI systems. This framework enables stakeholders to systematically compare various agents and identify the specific capabilities (e.g., advanced reasoning or continuous operation) necessary for particular use cases. In addition, it can help guide the development of novel agentic AI systems by clarifying which dimensions should be prioritized to meet defined design or operational goals. From a practical standpoint, the typology also aids organizations in determining whether a system's current level of autonomy or adaptiveness aligns with its broader strategic objectives. Finally, as AI technology progresses, the framework can be revised to incorporate new dimensions or categories, ensuring it remains a relevant and robust tool for evaluating and shaping the next generation of agentic AI solutions.

# Discussion and Conclusion

This paper sets out to understand the rapidly evolving "agentic turn" in AI research, from passive systems to those that sense, reason, decide, and act with minimal human supervision, allowing them to perform tasks once thought difficult to automate. We approached our research objective through the theoretical lens of agency, constructing a typology in six phases to classify the diverse capabilities of agentic AI systems. The resulting framework, built around eight dimensions, captures the progression from non-agentic to highly agentic systems. The typology offers a coherent foundation for evaluating and designing agentic AI with relevant contributions to research and practice by formalizing the core capabilities of knowledge scope, perception, reasoning, interactivity, reasoning, adaptability, interactivity, and operation. Further, drawing on empirical AI systems, we demonstrated how the typology can be applied and how a reduced version can be tailored to specific scenarios or use cases.

A key contribution of our work is a precise definition of an AI system across different sophistication levels. Grounded in four agency conditions, the typology can serve as a consistent boundary object for examining agency in AI. Each dimension isolates an observable, technology-agnostic capability, making the framework a versatile scaffold. By mapping systems against these dimensions, researchers can distinguish agentic from non-agentic variants, trace capability progressions, and study how differing "agentic profiles" interact with organizational routines. The typology can also act as a launchpad for theorizing application scenarios, downstream consequences, and design-science artifacts, among other directions. As AI evolves, the typology can be extended with new dimensions or categories, maintaining its relevance and generally serving as a tool for researching future trajectories of AI developments.

Additionally, the typology offers a decision-support tool for practitioners, for which we have demonstrated how the typology can be applied to real-world agentic AI systems. Mapping AI capabilities along standardized dimensions helps identify key differences between systems, facilitating informed decision-making when selecting and integrating agentic AI solutions and evaluating whether a candidate's characteristics and abilities fit a desired application scenario. In addition, it can help guide the development of novel agentic AI systems by clarifying which dimensions should be prioritized to meet defined design or operational goals. The typology also aids organizations in determining whether a system's current sophistication levels across the characteristics align with its broader strategic objectives and assists in preparing to facilitate a smooth transition from non-agentic systems.

Beyond its research and practical implications, the typology can act as an anchor point for potential consequences or risks. By decomposing agentic AI capabilities into clearly defined, measurable dimensions, the framework allows investigation of potential harms associated with increasingly agentic systems – such as shifts in decision-making authority away from humans, ethical dilemmas, or systemic risks such as workforce displacement (Chan et al., 2023) – thereby informing appropriate safeguards and governance mechanisms, both from an inter-organizational, societal, or policy perspective.

Despite its contributions, our approach has limitations. As with any qualitative analysis, there is inherent subjectivity in defining dimensions and characteristics, which may influence how agentic AI systems are interpreted and categorized. In the evaluation phase, which utilized the AI system DR, the quality of outputs was constrained by the availability and scope of the sources accessible to the model. While DR offered valuable insights, its reasoning relied on data from potentially incomplete or non-peer-reviewed sources, and due to the volume of references, we could not independently verify every citation it generated. We implemented multiple validation cycles to address this and maintained human oversight. Additionally, our construction process included arXiv publications frequently used in AI to capture recent developments. To ensure reliability, we followed established criteria for assessing the quality of non-peer-reviewed literature (Garousi et al., 2019). Finally, although the typology was designed to be agnostic, the prominence of LLM-based agentic systems may have introduced an unintended bias toward the types of agentic behavior these systems exhibit. We attempted to mitigate this by grounding the framework in canonical agency conditions and incorporating non-agentic and partially agentic systems; however, further empirical validation across a broader range of AI models will be necessary to refine and generalize the typology.

In summary, the typology contributes to understanding agentic AI systems in a rapidly evolving field, capturing recent developments and outlining potential future trajectories. It offers a valuable foundation for researchers and practitioners by enabling more informed decision-making about adopting or designing AI solutions. As a next step, the typology could be tested and refined through expert interviews and practical case studies, deepening our insight into how these frameworks are best applied in organizational contexts. Such further exploration would help validate and refine the constructs, ensuring the typology remains robust, relevant, and well-suited for guiding the deployment and evolution of agentic AI systems in real-world settings. Additionally, the typology can provide a foundation for research that explores the organizational changes, implementation strategies, and resource requirements needed for effective adoption. Further, outlining AI capabilities along a spectrum of agency offers a structured lens for examining potential outcomes, ethical concerns, policy implications, and harms associated with deploying agentic AI systems.